\documentclass[11pt]{article}
\usepackage{eacl2014}
\usepackage{times}
\usepackage{url}
\usepackage{latexsym}
\usepackage{graphicx}

\title{On Detecting Messaging Abuse in Short Text Messages using Linguistic and Behavioral patterns}
  
\author{Alejandro Mosquera \\
  Symantec Ltd. \\
   Reading, United Kingdom \\
  {\tt \tiny{alejandro\_mosquera@symantec.com}} \\\And
  Lamine Aouad \\
  Symantec Ltd. \\
  Dublin, Ireland \\
  {\tt \tiny{lamine\_aouad@symantec.com}} \\\And
  Slawomir Grzonkowski \\
  Symantec Ltd. \\
  Dublin, Ireland \\
  {\tt \tiny{slawomir\_grzonkowski@symantec.com}}  \\\And
  Dylan Morss \\
  Symantec Ltd. \\
  San Francisco, United States \\
  {\tt \tiny{dylan\_morss@symantec.com}}\\}
  
\date{}

\begin{document}
\maketitle
\begin{abstract}
The use of short text messages in social media and instant messaging has become a popular communication channel during the last years. This rising popularity has caused an increment in messaging threats such as spam, phishing or malware as well as other threats.
The processing of these short text message threats could pose additional challenges such as the presence of lexical variants, SMS-like contractions or advanced obfuscations which can degrade the performance of traditional filtering solutions. By using a real-world SMS data set from a large telecommunications operator from the US and a social media corpus, in this paper we analyze the effectiveness of machine learning filters based on linguistic and behavioral patterns in order to detect short text spam and abusive users in the network. We have also explored different ways to deal with short text message challenges such as tokenization and entity detection by using text normalization and substring clustering techniques. The obtained results show the validity of the proposed solution by enhancing baseline approaches.

\end{abstract}

\section{Introduction}
During the last years the use of short text messages in social media and instant messaging has become a powerful communication channel where users can exchange information, connect and share links with each other. As happens with another popular platforms such as email we have witnessed an increment on messaging threats such as spam (i.e., advertising or affiliate campaigns), scam (i.e., financial fraud), phishing (i.e., attacks with aim to steal sensitive information) and the spread of malicious software (i.e., worms, ransomware, spyware or Trojan horses) between others \cite{grzonkowski2014smartphone}. While these share common features with campaigns seen on email such as the use of an URL or phone number as a call-to-action, short text message spam usually lack server and metadata-rich information. Also messages are very brief due to size restriction to a few number of characters and do not always contain \textit{spammy} keywords e.g \textit{``This is the thing that I told u about http://xxxxx''}. Because of the real-time nature of these conversations and the need to process a high volume of data there are additional performance restrictions that can limit the analysis time. For example, SMS messages cannot be delayed in most of the cases for more than a few seconds, so taking into account that some campaigns are active just for some minutes a very fast response is needed in order to block these threats in time.  Because all these features short text message spam can be challenging for traditional filtering solutions \cite{Cormack2008}.  Also, the presence of obfuscations, lexical variants or wordplay and the absence of an explicit call-to-action can cause that the same short message can categorized as malicious or not depending on the context e.g. \textit{``I've met you last night call me back''}. This may degrade the performance of machine learning filters, requiring some degree of adaptation.

One of the biggest handicaps that the scientific community has faced regarding this topic is the limited availability of public short text message spam datasets mainly because of privacy and legal reasons. Another existing limitation is that these usually do not contain additional metadata such as sender and recipient identifiers, time stamp or network information. For this reason it is difficult to test and evaluate of possible filtering solutions without a real-world scenario.

Using a real-world Short Messaging Service (SMS) data set from a large telecommunications operator from the US, we analyze the effectiveness of machine learning filters based on linguistic patterns in order to detect and stop short text spam. Because textual features alone can generate false-positives in some situations such as forwarded messages or messages with ambiguous URLs, we evaluate the combination of behavioral and linguistic information in order to develop more robust models for detecting malicious users in the network. We also study the use of the same approaches for social media spam filtering (comment spam) and the contribution of text normalization techniques to the performance of the proposed solutions.

The paper is structured as follows: in Section 1 we show the related work, then in Section 2 our combined model for detecting spammers using linguistic and behavioral patters are explained, the used datasets are introduced in Section 3, the experimental results are described in Section 4, Section 5 contains the discussion of the results and in Section 4 we draw the conclusions and propose future work.

\section{Related Work}
Most of the first studies about short text message abuse are related with SMS spam where initial analyses focused on content analysis \cite{GomezHidalgo2006} based on n-gram and text count features \cite{Cormack2007}, \cite{Cormack2007b} using machine learning algorithms. The best classification results for SMS were obtained with support vector machines (SVM) with a 0.95 AUC.
 
Another content-based approaches for SMS spam used pattern matching \cite{Liu2010}, near-duplicate detection \cite{Coskun2012}, \cite{Valles2011}, byte-level features \cite{Rafique2010}, evolutionary algorithms \cite{Rafique2011} and feature selection based on information gain and chi-square metrics \cite{Uysal2012}.

About studies using an specific deployment platform we can highlight the work of Narayan and Saxena\shortcite{Narayan2013} where they analyzed the performance of spam filtering apps in Android and proposed a combination of Bayesian and SVM classifiers evaluated on a corpus of 700 messages.

Regarding non-content features the use of the grey phone space has been applied in order to detect spammers targeting randomly generated subscriber phone numbers \cite{Jiang2013}. Additional metadata such as sender location, network usage and call detail records have been shown useful for mining behavioral patters of SMS spammers \cite{Murynets2012}. Also, both sending and temporal features such as message and recipient counts per specified periods of time \cite{Wang2010}, \cite{Xu2012} can be used in order to detect abusive SMS senders in mobile networks using a probabilistic model.

Regarding the use of content and behavioral features on social networks for spam filtering Benevenuto et. al \shortcite{benevenuto2010detecting} proposed a machine learning model in order to detect spammers in Twitter with almost a 70\% success rate. Using features based on profile metadata Lee et. al \shortcite{Lee2010} studied the discovery of social media spam in Twitter and MySpace with a 0.88 and 0.99 F1 respectively in order to create a social honeypot. Using a different approach based on sending patterns and targeting strategies Youtube comment spam campaigns were tracked by using network analysis \cite{calla2012}.

Most of these works were evaluated on small text spam collections \cite{Almeida2011}, \cite{Hidalgo2012} which can not be always representative if we take into account the fast changing nature of short text message spam campaigns. Also, these static datasets do not contain information about senders, recipients, network or additional metadata such as URL information, which can be relevant for filtering purposes. For this reason, in this paper we explore the problem of short text spam filtering on a live SMS feed from a large telecommunications operator from the US and we evaluate the performance of different filtering techniques.

\section{Modelling Linguistic and Behavioral Patterns}

Machine learning is a branch of Artificial Intelligence (AI) that allow computers to learn patterns and take decisions based on data. Because machine learning has been proved successful for text categorization tasks \cite{Sebastiani2002} such as topic detection \cite{Ghiassi2012} or spam email filtering \cite{Gunal2006} supervised machine learning models have been trained in order to automatically classify SMS senders using the features extracted from the training data. Because fighting spam is a success story of real-world machine learning we have trained a combined model using random forests \cite{Breiman2001} based on two systems: one using lexical patterns and based on message linguistic analysis (MELA) and another leveraging behavioral information using messaging pattern analysis (MPA) in order to identify short text message spammers using mobile networks.

\subsection{Message Linguistic Analysis: MELA}
Short text message spam on platforms such as SMS, instant messaging (IM) or microblogs usually contain an URL or phone number as an explicit call-to-action (CTA) in order to get conversions and monetize the campaign e.g. \textit{``Your credit card has been BLOCKED please visit http://xxxxxx.xx to reactivate''}. But this CTA can be also implicit \textit{``Are you tired of money problems? reply for a payday loan today''} thus making traditional filtering techniques such as URL or phone number reputation less effective against these type of threats.

Previous approaches \cite{Cormack2007}, \cite{Cormack2007b} used traditional n-gram features based on shallow natural language processing (NLP) techniques such as tokenization or lemmatization. However, the word-vector space model (WVSM) can be quite sensible to outliers generated by new spam campaigns and it can generate models with a high dimensionality due to the presence of lexical variants and intentionally obfuscated elements when training on big datasets. We have identified several spam campaigns showing some of these features thus making standard tokenization and entity detection techniques useless (see Table \ref{table:table7}).

\begin{table*}[t]
\begin{small}
\begin{center}
\begin{tabular}{|l|}
\hline
 Examples of incorrectly processed messages\\
\hline 
    \textbf{Message}: \textbackslash{}rOtOcarr0.K0nprare/0ld/trashed/crashed up/k.a.r.s/\textbackslash{}callus now555O5O5O5O\\
    \textbf{Tokens}: ['rOtOcarr0', 'K0nprare', '0ld', 'trashed', 'crashed', 'up', 'k', 'a', 'r', 's', 'callus', 'now555O5O5O5O']\\ \textbf{CTA}:[none]\\
    \textbf{Problem}: Incorrect tokenization and no CTA detection\\
    \hline 
     \textbf{Message}: Hi, I want to meet you tonight, spamdomain.com.Support me\\
     \textbf{Tokens}: ['Hi', 'I', 'want', 'to', 'meet', 'you', 'tonight', 'spamdomain.com.Support', 'me'\\
      \textbf{CTA}:['spamdomain.com.Support']\\
      \textbf{Problem}: Incorrect URL detection (wrong TLD)\\
     \hline 
     \textbf{Message}: Ive had a lot of fun tonight.tk\\
     \textbf{Tokens}: ['Ive', 'had', 'a', 'lot', 'of', 'fun', 'tonight.tk']\\
      \textbf{CTA}:['tonight.tk']\\
      \textbf{Problem}: Incorrect URL detection (no URL)\\
\hline
\end{tabular}
\caption {Examples of standard tokenization and CTA detection on obfuscated messages.}
\label{table:table7}
\end{center}
\end{small}
\end{table*}		   

For this reason and also because the relatively high amount of training data that we were using in comparison with state of the art approaches for SMS spam (600k messages) we decided to do not use a word vector space model. Instead we have clustered substring tokens from a subset of 100k messages using t-distributed stochastic neighbor embeddings (t-SNE) \cite{maaten2008visualizing}, string similarity functions based on matching n-grams and word co-occurrences. These substring tokens were mined from spam messages by taking into account the longest common substrings (LCS) of the most frequent words in the dataset after removing stopwords. The resulting 22 substring clusters were obtained after manually pruning the embeddings in order to remove bad-quality or non-relevant groups (see Table \ref{table:table1}).
Because the languages of these initial subset of SMS messages were mostly in English (95\%) and Spanish (2\%), the extracted substrings will reflect that lexical variety.

\begin{table*}[t]
\begin{small}
\begin{center}
\begin{tabular}{|l|l|}
\hline
 Cluster&Substrings\\
\hline
car&[auto, scrap, car, automob, coche, recycl]\\
job&[work, job, home, employ, paid, hire, salary, trabaj, salario, casa, income]\\
alarm&[alarm, alert, attention, danger, urgen, emergency]\\
event&[club, ticket, party, dj, gig, concert, artist, venue, disco, bar, lounge, club, fiesta]\\
phishing&[faceb, twitter, microsoft, paypa, twit, verizon, bank, visa, tweet, vzw, wmart, banco]\\
scam&[craig, verif, secur, cnbc, stock, safe, trade, coin, bonus, bill, card, busine, forex, broker, biz, ebook, rich, sell]\\
intent&[participa, redeem, choose, pick, order, llama, envia, enrol, enter, subscribe, dial, end, stop, register, read, get...]\\
bank&[invest, wire, limit, flag, master, debit, credit, visa, account, hack, activ, block, verify, compromise, disable...]\\
money&[cash, dollar, pound, euro, money, dinero, dolar, pesos, rupee]\\
...&[...]\\
\hline
\end{tabular}
\caption {Subset of MELA substring clusters for SMS spam.}
\label{table:table1}
\end{center}
\end{small}
\end{table*}

These substrings will generate vector counts for each cluster by using the Aho-Corasick algorithm \cite{aho1975}. 
Entities such as URLs, emails and phone numbers are identified and extracted. In the case of URLs these are processed in a special way as these are one of the preferred CTA vectors used by spammers (and in some cases the short message will contain just an URL) with more than a 70\% of the spam messages in our collected data. An additional feature vector will be generated for each URL and it will be processed by a separate classifier, converting the output of these results into MELA features (DOMAIN\_MELASCORE).

After analyzing textual patterns of short text spam messages we have identified 51 linguistic features that can be grouped into three different categories by taking into account their nature: Entity-based, Heuristic and Structural (see Table \ref{table:table2}).

\subsubsection{Entity Features} 
\begin{figure}[ht!]
\centering
\includegraphics[width=70mm]{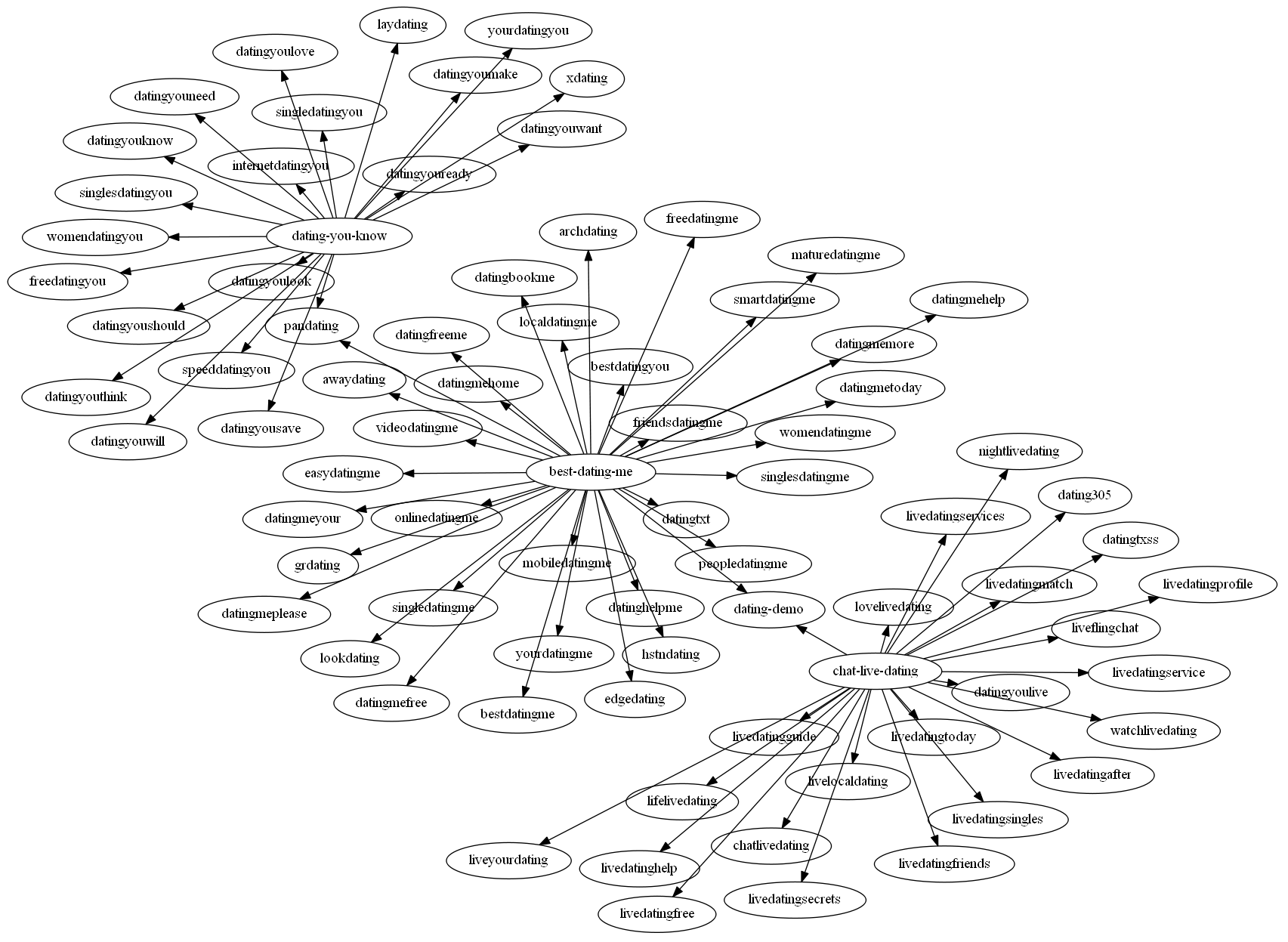}
\caption{Example of n-gram clustering for domain names serving adult affiliate campaigns.}
\label{figure:figure5}
\end{figure}

Besides basic entity counts (NUM\_OF\_URLS, NUM\_OF\_PHONES and NUM\_OF\_EMAILS) we have also identified temporal expressions (NUM\_OF\_TIMEX), numbers (NUM\_OF\_NUMBER) and currency mentions (NUM\_OF\_CURRENCY). Because URLs and especially domain names registered by spammers usually contain semantic and lexical patterns, we have processed them individually by using a subset of MELA features plus ad-hoc heuristics. We have observed that domain names in English serving the same campaigns had a high overlap of substring n-grams, which those can be characterized by using the previously obtained substring clusters (see Figure \ref{figure:figure5}).

\subsubsection{Heuristic Features}
Several heuristic features has been added in order to detect linguistic patterns commonly seen in short text spam campaign (ENDS\_WITH\_NUM, CONTAINS\_YEAR...) to discover Twitter hashtags and mentions (HEUR\_TWEET) or to identify shortening services (DOMAIN\_ISSHORT), commonly exploited free TLDs such as .tk or .pw that are abused by spammers (BADTLDS, SUSPTLDS, NORMALTLDS).

\subsubsection{Structural Features}
\begin{figure*}[ht!]
\centering
\includegraphics[width=140mm]{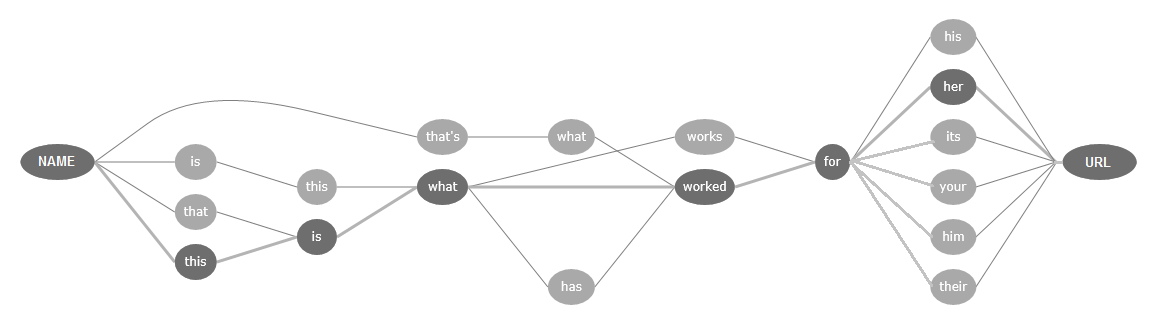}
\caption{Example of lexical variation and templates used by SMS spammers.}
\label{figure:figure4}
\end{figure*}

We observed campaigns that were following the same, but slightly modified, structure with a goal of evading detection. In order to do this they generate textual variants of the same message using randomized templates: e.g. \textit{``Andrea, this is what worked for his http://xxxxxxx''}, \textit{``Ashley, thats what works for you http://xxxxxxx''} (see Figure \ref{figure:figure4}). For this reason we have encoded the position of different CTA entities (PHONE\_POS, EMAIL\_POS, NUMBER\_POS) as a numeric vector by taking into account if these are at the beginning, at the end or in the middle of the message.

\begin{table}
\begin{small}
\begin{center}
\begin{tabular}{|l|l|}
\hline
 Message Features& Domain Features\\
\hline 
NUM\_OF\_URLS & STARTS\_WITH\_NUM\\
    NUM\_OF\_PHONES & ENDS\_WITH\_NUM\\
    NUM\_OF\_EMAILS& CONTAINS\_00\\
    URL\_POS & CONTAINS\_VV\\
    PHONE\_POS & CONTAINS\_YEAR\\
    EMAIL\_POS & CONTAINS\_1\\
    NUMBER\_POS & CONTAINS\_ZERO\\
    CONTAINS\_FWD & DIGIT\_RATIO\\
    LENGTH & HYPHEN\_COUNT\\
    WORD\_COUNT & LENGTH\\
    PHONEME\_COUNT & WORD\_COUNT\\
    SUBSTRING\_CLUST\_0 & PHONEME\_COUNT\\
     ... & SUBSTRING\_CLUST\_0\\
    SUBSTRING\_CLUST\_21 & ...\\
    UNSUBSCRIBE & SUBSTRING\_CLUST\_21\\
    PHONE\_ISFREE & CONTAINSWWW\\
    EMAIL\_ISFREE & BADTLDS\\
    URL\_ISDOM & SUSPTLDS\\
    DOMAIN\_MELASCORE & NORMALTLDS\\
    DOMAIN\_ISSHORT & ISSHORT\\
    NGRAM\_ENTROPY &\\
    START\_WITHNUMBER&\\
    END\_WITHNUMBER&\\
    TOKEN\_RATIO&\\
    NUM\_OF\_TIMEX&\\
    NUM\_OF\_NUMBER&\\
    NUM\_OF\_CURRENCY&\\
    STARTSWITH\_HELLO&\\
    ENDSWITH\_CTA & \\
    DOMAIN\_OBFUSCATION &\\
    HEUR\_TWEET &\\
\hline
\end{tabular}
\caption {MELA domain and message features.}
\label{table:table2}
\end{center}
\end{small}
\end{table}		   

\subsubsection{Text Normalization}
The language used in SMS messages usually can contain informal elements such as misspellings, slang, lexical variants and inconsistent punctuation, contractions, phonetic substitutions or emoticons. Spammers are aware of this peculiarities and generate messages that are very similar to ham by using creative variants of the original content in order to avoid fingerprinting. Text normalization techniques deal with out-of-vocabulary words (OOV) by replacing these variants with canonical ones \cite{aw2006}.
We have applied a text normalization process based on the exception dictionaries of TENOR \cite{mosquera2012towards} in order to substitute the most common English shortenings and lexical variants.

\subsection{Messaging Pattern Analysis: MPA}
While some spammers prefer to send thousands of messages in a short amount of time, which can raise volumetric alerts and get them blocked automatically, others prefer to go low volume or split the sending process over thousands of unique senders \cite{aouad2014sms}. One of the keys of successful spam campaigns is avoiding early detection, so reusing previous messages or senders will not be effective if the network has some basic level of anti-spam protection e.g. sender/CTA reputation or a hash database of spam messages. For this reason, we have extracted messaging features based on sender behavior in order to detect abusive activity in the network by collecting information from senders which at least send 50 messages in a 7-day period, these values are empiric and can be changed depending on the network and platform. In order to do this we have collected communication-specific metadata such as origin and destination network and if these are US-based or not (ORIG\_NETWORK, DEST\_NETWORK, SENDER\_NETWORK\_IS\_NOT\_US, DEST\_NETWORK\_IS\_NOT\_US, NUM\_OF\_UNIQUE\_DEST\_NETWORKS). Because spammers can target individual users leaked from contact information databases or randomly/uniformly-generated victims we model also their targeting strategy. For this reason, numeric features such as the number of sent messages per second that differentiate between slow and fast senders or sender number entropy, provides information about target randomness (NUM\_OF\_UNIQUE\_RECIPIENTS, SENDING\_FREQUENCY, RECIPIENT\_NUMBER\_ENTROPY). Also, all the MELA features for the first sent message are also included as part of MPA (see Table \ref{table:table3}) in order to cover both messaging and linguistic patterns.

\begin{table}
\begin{small}
\begin{center}
\begin{tabular}{|l|}
\hline
 MPA Features\\
\hline
 ORIG\_NETWORK \\
 DEST\_NETWORK \\
 SENDER\_NETWORK\_IS\_NOT\_US \\
 DEST\_NETWORK\_IS\_NOT\_US \\
 NUM\_OF\_UNIQUE\_RECIPIENTS \\
 RECIPIENT\_NUMBER\_ENTROPY \\
 NUM\_OF\_UNIQUE\_DEST\_NETWORKS \\
 SENDING\_FREQUENCY \\
 NUM\_OF\_UNIQUE\_MESSAGES \\
 MELA\_FEATURE\_0 \\
 ... \\
 MELA\_FEATURE\_50 \\
\hline
\end{tabular}
\caption {MPA features.}
\label{table:table3}
\end{center}
\end{small}
\end{table}	

\begin{figure}[ht!]
\centering
\includegraphics[width=65mm]{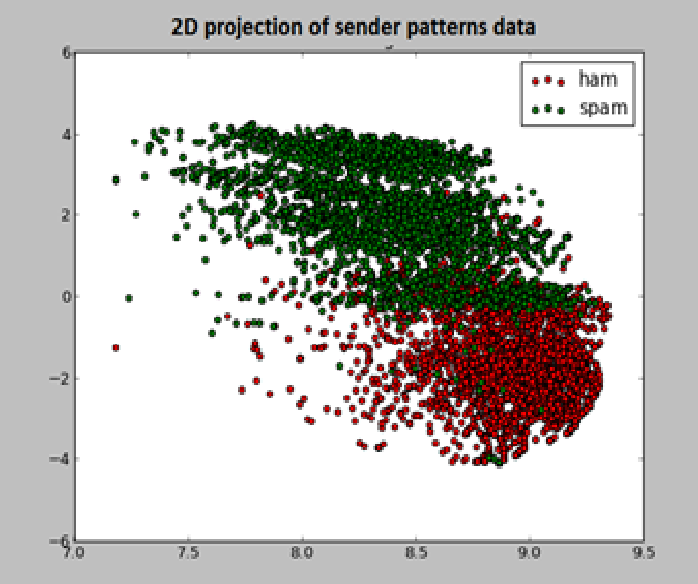}
\caption{2D projection of MPA sender patterns data.}
\label{figure:figure1}
\end{figure}

In Figure \ref{figure:figure1}, we can observe how the 2D projection of MPA features show clear differences between legit and spammer messaging patterns that are almost linearly separable, which depicts the accuracy of the engineered features.

\section{Datasets}
Previous studies on SMS spam used the datasets published by \cite{Almeida2011} and \cite{Delany2012}. Since all these were constructed using the NUS SMS corpus \cite{Chen2011} and Grumbletext\footnote{http://grumbletext.co.uk} after removing the duplicates the total number of messages result in less than 2500 tagged instances\footnote{https://github.com/okkhoy/SpamSMSData} by taking into account train and test sets. Because of the changing nature of SMS spam campaigns a bigger dataset is needed in order to create robust models.

For this reason, we have captured 300K unique spam and 100K ham messages from a large US telecommunications operator covering the period from April 2013 to December 2013. Because the number of unique ham messages were considerably smaller we have balanced the corpus by adding 200K tweets after removing all hash tags and user mentions in order to simulate clean SMS data. In addition we have compiled data from 40K senders with 50 sent SMS each in a week period (20K spammers and 20K legit).

We have also experimented with short message spam from social media, in order to do this a comment spam dataset \cite{calla2012} containing 6.4M comments with 481K marked as spam by users\footnote{http://mlg.ucd.ie/files/datasets/youtube\_comments\_20120117.csv.bz2} has been used.

\section{Results and Evaluation}

The proposed filtering models have been evaluated in terms of precision, recall and the harmonic mean of these (F1) by using a ten-fold cross validation with the datasets described in the previous section. Two baseline systems for evaluation have been developed by using common features such as word n-grams and sparse orthogonal n-grams (n=3).

For the SMS corpus on the one hand MELA results shows a 0.05\% false positive (FP) rate and a 9.4\% false negative (FN) rate for domain classification (see Table \ref{table:table6}). On the other hand, results for message classification show a 0.02\% false positive (FP) rate and a 3.1\% false negative (FN) rate (see Table \ref{table:table4}). MELA scored more than 5 F1 more points than the baseline. We can observe how the use of text normalization techniques can improve both n-gram and sgram models by almost 2 F1 points.

The results obtained after analyzing the Youtube comment spam corpus were considerably lower not only for MELA but also for baseline approaches with a 72\% F1. This means that lexical patterns are less relevant on these messages as many of them lack of URLs and explicit CTAs.

\begin{table}
\begin{center}
\begin{tabular}{|l|l|l|l|}
\hline
 &Precision&Recall&F1\\
\hline
Ham&0.73&1&0.84\\
Spam&1&0.62&0.77\\
avg.&0.86&0.81&0.81\\
\hline
\end{tabular}
\caption {MELA domain classification results for SMS spam.}
\label{table:table6}
\end{center}
\end{table}

\begin{table}
\begin{center}
\begin{tabular}{|l|l|l|l|l|}
\hline
Features&Class &Precision&Recall&F1\\
\hline
&Ham&\textbf{0.94}&1&\textbf{0.97}\\
MELA&Spam&1&\textbf{0.93}&\textbf{0.97}\\
&avg.&\textbf{0.97}&\textbf{0.97}&\textbf{0.97}\\
\hline
&Ham&0.83&1&0.91\\
SGRAM&Spam&1&0.79&0.88\\
&avg.&0.91&0.90&0.89\\
\hline
&Ham&0.86&1&0.92\\
SGRAM&Spam&1&0.84&0.91\\
NORM.&avg.&0.93&0.92&0.92\\
\hline
&Ham&0.81&1&0.89\\
NGRAM&Spam&1&0.75&0.86\\
&avg.&0.90&0.88&0.88\\
\hline
&Ham&0.82&1&0.90\\
NGRAM&Spam&1&0.79&0.88\\
NORM.&avg.&0.91&0.89&0.89\\
\hline
\end{tabular}
\caption {Message classification results for SMS spam.}
\label{table:table4}
\end{center}
\end{table}

\begin{table}
\begin{center}
\begin{tabular}{|l|l|l|l|l|}
\hline
Features&Class &Precision&Recall&F1\\
\hline
&Ham&\textbf{0.66}&0.99&\textbf{0.79}\\
MELA&Spam&0.98&\textbf{0.48}&\textbf{0.64}\\
&avg.&\textbf{0.82}&\textbf{0.73}&\textbf{0.72}\\
\hline
&Ham&0.65&1&0.79\\
SGRAM&Spam&0.99&0.47&0.63\\
&avg.&0.82&0.73&0.71\\
\hline
&Ham&0.65&1&0.79\\
SGRAM&Spam&0.99&0.47&0.64\\
NORM.&avg.&0.82&0.73&0.71\\
\hline
&Ham&0.63&1&0.77\\
NGRAM&Spam&0.99&0.42&0.59\\
&avg.&0.81&0.71&0.69\\
\hline
&Ham&0.64&1&0.78\\
NGRAM&Spam&0.99&0.45&0.62\\
NORM.&avg.&0.82&0.72&0.70\\
\hline
\end{tabular}
\caption {Message classification results for Youtube comment spam.}
\label{table:table6}
\end{center}
\end{table}

MPA results were similar for sender classification with a 0.08\% and 3\% FP and FN rates respectively (see Table \ref{table:table5}). 
Regarding the machine learning hyperparameters we have noticed that increasing the number of trees had a positive impact in the results, finding that n=500 was the optimal value. No relevant improvements were found when using a higher number of estimators.

\begin{table}
\begin{center}
\begin{tabular}{|l|l|l|l|}
\hline
 &Precision&Recall&F1\\
\hline
Ham&0.95&1&0.98\\
Spam&1&0.95&0.97\\
avg.&0.98&0.97&0.97\\
\hline
\end{tabular}
\caption {MPA sender classification results for SMS spam.}
\label{table:table5}
\end{center}
\end{table}

\begin{figure}[ht!]
\centering
\includegraphics[width=65mm]{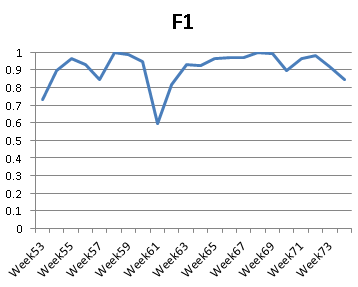}
\caption{F1 score evolution for sender classification of live SMS data during 22 weeks using the initial MPA model.}
\label{figure:figure2}
\end{figure}

\begin{figure}[ht!]
\centering
\includegraphics[width=65mm]{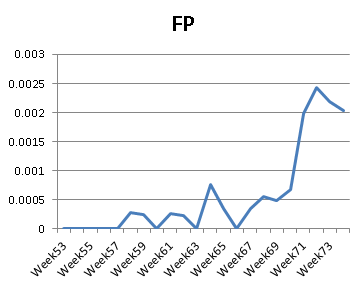}
\caption{False positive rate evolution for sender classification of live SMS data during 22 weeks using the initial MPA model.}
\label{figure:figure3}
\end{figure}

We have also evaluated the trained MPA model against live SMS data from the same US telecom operator for a 22-week period in terms of F1 and FP rates with an average 91\% and 0.058\% respectively (see Figure \ref{figure:figure2} and Figure \ref{figure:figure3}). Taking into account that the models have not been retrained during the whole evaluation period these results are quite competitive in comparison with the ones obtained by cross validation. Spam campaigns evolve quickly over time \cite{aouad2014sms} and new types of messaging threats appear every day effectively degrading MPA performance over the time in terms of FP's.

MPA classification F1 remained more or less constant during the whole period with the exception of the spike in the chart on week 62 to 63. There were a high number of false negative senders from an adult scam campaign targeting chat users: \textit{``hey babe saw youre pic online lets chat on mssnger add me xxxxxx''} during those weeks, affecting the overall efficacy. This campaign used an implicit CTA and showed a targeted-attack pattern using recipients extracted from social media and adult dating websites. Because the number of non-explicit CTA campaigns in the training data was small the campaign was missed. However these could be detected by adding them to the dataset and retraining the model.

\section{Discussion}

As mentioned in the previous section, the absence of an explicit CTA or the use of URL shortening services are some of the challenges found during the analysis of the obtained results. Because these are present in the training data and can generate a high number of false positives a cost-sensitive classification was used in order to effectively lowering the number of FP's but also compromising spammer detection efficacy. We have also showed how the use of simple text normalization and substring clusters can generate efficient and agile models suitable for real-time filtering. About the first, it provided a positive but modest contribution by pushing the F1 of message classification further.  There is still room for improvements on this area that will require a detailed case-by-case analysis that we address in our future work. Regarding the second there are obvious advantages over standard tokenization approaches but they also can generate false positives, which will require further work in order to identify potentially problematic cases. 

\section{Conclusions and Future Work}

In this paper we have explored the use of lexical and behavioral patterns in order to detect threats in short text message networks. Two different filtering systems have been proposed for message filtering and abusive sender identification. These were evaluated using both a relatively-big static corpus and live network data for a 22-week period. The obtained results show the validity of the proposed solution by enhancing baseline approaches. We have also explored different ways to deal with short text message challenges such as tokenization and entity detection by using text normalization and substring clustering techniques. The use of these was found to not only slightly improve the proposed baselines but to be as well a more performance-wise solution. 

We have identified additional data sources that can be leveraged in order to improve the obtained results such as the use of URL content and metadata, URL reputation databases and WHOIS domain information. The use of features based on these and the generation of fingerprints for detected campaigns are left to a future work.

\bibliographystyle{acl}
\bibliography{paper}

\end{document}